\documentclass[letterpaper]{article}
\usepackage[utf8]{inputenc}
\usepackage[T1]{fontenc}

\usepackage{custom}
\usepackage{times}
\usepackage{helvet}
\usepackage{courier}
\usepackage{epigraph}

\usepackage{url}
\usepackage{breakurl}
\usepackage[hyperindex,breaklinks]{hyperref}

\usepackage[backend=bibtex, isbn=false, doi=false, sorting=none, natbib]{biblatex}
\renewbibmacro{in:}{}

\bibliography{article}

\begin{document}
%
\title{Building Safer AGI by introducing Artificial Stupidity}
\author{Michaël Trazzi$^1$, Roman V. Yampolskiy$^2$\\
$^1$Sorbonne universite, Paris, France \\
$^2$Computer Engineering and Computer Science, Speed School of Engineering, University of Louisville \\
\texttt{michael.trazzi@etu.upmc.fr} \\
\texttt{roman.yampolskiy@louisville.edu} \\
}

\maketitle

\begin{abstract}
\begin{quote}
Artificial Intelligence (AI) achieved super-human performance in a broad variety of domains. We say that an AI is made \textit{Artificially Stupid} on a task when some limitations are deliberately introduced to match a human's ability to do the task. An Artificial General Intelligence (AGI) can be made safer by limiting its computing power and memory, or by introducing Artificial Stupidity on certain tasks. We survey human intellectual limits and give recommendations for which limits to implement in order to build a safe AGI.
\end{quote}
\end{abstract}

\section*{Introduction}

\noindent The \textit{Turing Test} \cite{Turing1950} was designed to replace the philosophical question "Can machines think ?" with a more pragmatic one: "Can digital computers imitate human behaviors in answering text questions ?" Since 1990, the \textit{Loebner Prize} competition awards the AI that has the more human-like behaviour when passing a five-minutes Turing Test. This contest is controversial within the field of AI for encouraging low-quality interactions and simplistic chatbots \cite{Shieber1994}. 

Programmers force chatbots to make mistakes, such as typing errors, to be more human-like. Because computers achieve super-human performance in some tasks, such as arithmetic \cite{Turing1950} or video game \cite{Liden2003}, their ability may need to be artificially constrained. Those deliberate mistakes are called \textit{Artificial Stupidity} in the media \cite{Economist1992} \cite{Salon12003}.

Solving the Turing Test problem in its general form implies building an AI capable of delivering human-like replies to any question of human interest. As such, it is known to be AI-Complete \cite{Yampolskiy2013}, because by solving this problem we would be able to solve most of problems of interest in Artificial Intelligence by reformulating them as questions during a Turing Test. To appear human, an AI will need to fully understand human limits and biases. Thus, the Turing Test can be used to test if an AI is capable of understanding \textit{human stupidity}.

By deliberately limiting an AI's ability to achieve a task, to better match  humans' ability, an AI can be made safer, in the sense that its capabilities will not exceed by several orders of magnitude humans' abilities. Upper-bounds on the number of operations made per second by a human brain have been estimated \cite{Moravec1997} \cite{Bostrom1998}. To obtain an AI that does not exceed by far humans' abilities, for instance in arithmetics, the computing power allowed for mathematical capabilities must be artificially diminished. Besides, humans exhibit cognitive biases, which result in systematic errors in judgment and decision making \cite{Haselton2005}. In order to build a safe AGI, some of those biases may need to be replicated.

We will start by introducing the concept of Artificial Stupidity. Then, we will recommend limitations to build a safer AGI. 

\section{Artificial Stupidity} 
\subsection{Passing the Turing Test}

To pass the Turing Test, programs that are deliberately simplistic perform better at Turing Test contests such as the Loebner Prize. The computer program A.L.I.C.E. (or Artificial Linguistic Internet Computer Entity) \cite{Wallace2009} won the Loebner Prize in 2000, 2001 and 2004, even though "there is no representation of knowledge, no common-sense reasoning, no inference engine to mimic human thought. Just a very long list of canned answers, from which it picks the best option" \cite{Salon22003}. A.L.I.C.E. has an approach similar to ELIZA \cite{Weizenbaum1966}: it identifies some relevant keywords and give appropriate answers without learning anything about the interrogator \cite{Salon22003}.

A general trend for computer programs written for the Loebner prize is to avoid being asked questions it cannot answer, by directing the conversation towards simpler conversational context. For A.L.I.C.E. and ELIZA, that means focusing mainly of what had been said in the last few sentences (stateless context). Another example of an AI performing well at Turing Test contests is Eugene Goostman, who convinced 33\%  of the judges that it was human \cite{Zdnet2014}. Goostman is portrayed as a thirteen-year-old Ukrainian boy who does not speak English well. Thus, Goostman makes typing mistakes and its interrogators are more inclined to forgive its grammatical errors or lack of general knowledge. Introducing deliberate mistakes, what we call \textit{Artificial Stupidity}, is necessary to cover up an even greater gap in intelligence during a Turing Test.

\subsection{Interacting with Humans}

Outside those very specific Turing Test contests, Artificial Stupidity is being increasingly introduced to interact with humans. In “Artificial  Stupidity:  The  Art  of  Intentional Mistakes”\cite{Liden2003}, Liden describes the design choices an AI programmer must make in the context of video games. He gives general principles that a Non-Player Character (NPC) must follow to make the game playable. For instance, NPCs must "move before firing" (e.g. by rolling when the player enters the room) so that the player has additional time to understand that a fight is happening, or "miss the first time" to indicate the direction of the attack without hurting the player. In video games, because computer programs can be much more capable than human beings (for instance because of their perfect aim in First Person Shooters (FPS)), developers force NPCs to make mistakes to make life easier for the human player.

This tendency to make AI deliberately stupid can be observed across multiple domains. For example, at Google I/O 2018, Sundar Pichai introduced Google Duplex, "a new technology for conducting natural conversations to carry out “real world” tasks over the phone"\cite{Duplex2018}. In the demo, the Google Assistant used this Google Duplex technology to successfully make an appointment with a human. To that end, it used the interjection "uh" to imitate the space filler humans use in day-to-day conversations. This interjection was not  \textit{necessary} to make the appointment. More precisely, when humans use these kind of space fillers, it is a sign of poor communication skills.  The developers introduced this type of Artificial Stupidity to make the call more fluid, more human-like.

\subsection{Exploiting Human Vulnerabilities}

When interacting with AI, humans want to fulfill some basic needs and desires. The AI can exploit these cravings, by giving people what they want. Accordingly, the AI may want to appear vulnerable to make the human feel better about himself. For instance, Liden \cite{Liden2003} suggests to give NPCs "horrible aim", so that humans feel superior because they think they are dodging gunshots. Similarly, he encourages "kung-fu style" fights, where amongst dozens of enemies, only two are effectively attacking the player at each moment. Thus, the player feels powerful (because he believes he is fighting multiple enemies). Instead of unleashing its full potentiality, the AI designers diminish its power to please the players.

The same mechanism applies to the computer program A.L.I.C.E.. The program delivers a simple \textit{verbal behavior}, and because "a great deal of our interactions with others involves verbal behavior, and many people are interested in what happens when you talk to someone"\cite{Salon22003}, it fulfills this elementary craving of obtaining a verbal behavioral response. In his "Artifical Stupidity" article, part 2, Sundman quotes Wallace, the inventor of A.L.I.C.E, explaining why he thinks humans enjoy talking to his program: "It's merely a machine designed to formulate answers that will keep you talking. And this strategy works, [...] because that's what people are: mindless robots who don't listen to each other but merely regurgitate canned answers."

More generally, virtual assistants and chatbots are technologies aimed at helping consumers. To that end, they must provide value. They might achieve that by gaining users' thrust and improving their overall well-being. For instance, Woebot, a Facebook chatbot developed by Stanford Researchers, improves users' mental-health through Cognitive Behavorial Therapy (CBT) \cite{MIT2017}. In addition to CBT, Woebot uses two therapeutic process-oriented features. First, it provides empathic responses to the user, according to the mood the user said he had. Second, content tailored to its mood is presented to the user. \cite{Woebot2017}. The chatbot delivers specifically what will help the user best, without describing the details of the user's mental-health. It gives a very simple answer the consumer wanted, masking the intelligence of its algorithm. Chatbots are designed to help humans, giving appropriate simple responses. They are not designated to appear smarter than humans.

\subsection{Avoiding Superintelligence requires Superintelligence}

In "Computing Machinery and Intelligence" \cite{Turing1950}, Turing exposes common fallacies when arguing that a machine cannot pass the Turing Test. In particular, he explains why the belief that "the interrogator could distinguish the machine from the man simply by setting them a number of problems in arithmetic" because "the machine would be unmasked because of its deadly accuracy" is false. Indeed, the machine "would not attempt to give the right answers to the arithmetic problems. It would deliberately introduce mistakes in a manner calculated to confuse the interrogator." Thus, the machine would hide its super-human ability by giving a wrong answer, or simply saying that he could not compute it. Similarly, in a video game, AI designers artificially make the AI not omniscient, so that it does not miraculously guess where each and every weapon of the game is located \cite{Liden2003}.

The general trend here is that AI tend to quickly achieve super-human level performance after having achieved human-level performance. For instance, for the game of Go, in a few months, the state-of-the-art went from strong amateur, to weak professional player, to super-human performance \cite{Silver2016}. From that point onwards, to make the AI pass a Turing Test, or make it behave human-like to satisfy human desires, AI designers must deliberately limit its capabilities.

We now take the point of view of algorithmic complexity. We call \textit{AI-problems} the problems that can be solved (i.e. for which the correct output for a given input can be computed) by the union of all humans \cite{Yampolskiy2013}. The Turing Test is said to be \textit{AI-complete} \cite{Yampolskiy2013}, because all AI-problems can be reduced via a polynomial-time reduction to it (by framing the problem as a question during a Turing Test), and it is an AI-problem. Additionally,  we say that a problem is \textit{AI-Hard} if we can find a polynomial-time reduction from an AI-complete problem to this problem. In this setting, the problem of comprehensively specifying the limits of the human cognition to an AI is AI-Hard. Indeed, it implies being capable to know which questions a human can answer in a given time during a Turing Test, and which answers are likely to be produced by humans. Therefore, the Turing Test can be reduced polynomially to specifying the limits of human cognition, so specifying such limits is AI-Hard.

Although the precise limits of human cognition are not fully known, specific recommendations on minima or maxima for different capabilities can be given.

\section{The Cognitive Limits of the Human Brain}

\epigraph{I believe that in about fifty years time it will be possible to programme computers, with a storage capacity of about $10^9$, to make them play the imitation game so well that an average interrogator will not have more than 70 percent chance of making the right identification after five minutes of questioning.}{Turing, Computing Machinery and Intelligence, 1950}

Sixty-eight years ago, Turing estimated that in the 2000s, an AI with only  1 Gb in storage could pass a five minutes Turing Test 30\% of the time. Previously, we showed that passing the Turing Test \textit{in general} was AI-complete. The amount of computing resources necessary to pass the Turing Test is then a relevant estimate for determining the computing power necessary to attain Human-Level Machine Intelligence (HLMI). In what follows, we try to estimate the computing power of the human brain.

\subsection{Limits in Computing}

The brain is a complex system with an architecture completely different from the usual von Neumann computer architecture. However, estimates about the storage capacity of the brain and the number of operations per second were attempted.

\subsubsection{Long-term Memory}

Here is how Turing \cite{Turing1950} justifies the $10^9$ bits of storage capacity:

\begin{quote}
``Estimates of the storage capacity of the brain vary from $10^{10}$
to $10^{15}$ binary digits. I incline to the lower values and believe
that only a very small fraction is used for the higher types of
thinking. Most of it is probably used for the retention of visual
impressions.''
\end{quote}

The storage capacity of the brain is generally considered to be within the bounds given by Turing (resp. $10^{10}$ and $10^{15}$). Although the encoding of information in our brains is different from the encoding in a computer, we observe many similarities \cite{Lecoq2013}.

To estimate the storage capacity of the human brain, we first evaluate the number of synapses available in the brain. The human brain has about 100 billion neurons \cite{Williams1988}. Each neuron has about five thousand potential synapses, so this amounts to about $5.10^{14}$ synapses \cite{Bostrom1998}, so $5.10^{14}$ potential datapoints.

This shows that the brain could in theory encode between $10^{12}$ and $10^{15}$ bits of information, assuming that each synapses stores one bit. However, such estimates are still approximate because neuroscientists do not know precisely how synapses actually encode information: some of them can encode multiple bits by transmitting different strengths, and individual synapses are not completely independent \cite{Wickman2012}.

\subsubsection{Processing}

Even though the brain can encode Terabits of information, humans are in practice very limited in the amount of information they can process.

In his classical article \cite{Miller1956}, Miller showed how our minds could only hold about $7\pm2$ concepts in our working memory. More generally, three essential bottlenecks were shown to limit information processes in the brain: the Attentional Blink (AB) limits our ability to consciously perceive, the Visual Short-Term Memory (VSTM) our capacity to hold in mind, and the Psychological Refractory Period (RFP) our ability to act upon the visual world \cite{Marois2005}. In particular, the brain takes up to 100ms to process complex images \cite{Rousselet2004}.

Moreover, the processing time seems to take longer when the choice to make takes as input a complex information. This is known as \textit{Hick's Law} \cite{Hick1952}: the time it takes to make a choice is linearly related to the entropy  of the possible alternatives.

\subsubsection{Computing}

One approach to evaluate the complexity of the processes happening in the brain is to estimate the maximum number of operations per second. Thus, Moravec \cite{Moravec1997} estimates that to replicate all human's function as a whole one would need about 100 MIPS (Millions of Instructions per Second), by comparing it to the computational needs for edge extraction in robotics. Using the same estimation for the number of synapses in the brain (mentioned in \textit{Memory}), Bostrom \cite{Bostrom1998} concludes that the brain uses at most about $10^{17}$ operations per second (for a survey of the different estimates of the computational capacity of the brain, see (Bostrom, 2008) \cite{Bostrom2008}).

\subsubsection{Clock Speed}

The brain does not operate with a central clock. That's why the term "clock-speed" does not describe accurately processes happening in the brain. However, it if possible to compare the transmission of information in the brain and inside a computer.

Processes emerge and dissolve in parallel in different parts of the brain at different frequency bands: theta(5-8Hz), alpha(9-12Hz), beta(14-28Hz) and gamma(40-80Hz). Comparing computer and brain frequences, Bostrom notes that ``biological neurons operate at a peak speed of about 200 Hz, a full seven orders of magnitude slower than a modern microprocessor (\textasciitilde2 GHz)'' \cite{Bostrom2014}.

It is important to note that clock speed, alone, do not fully characterize the performances of a processor \cite{Smith2002}. Furthermore, the processes happening in the brain use several orders of magnitude more parallelization than modern processors.

\subsection{Cognitive biases}

\epigraph{Humans, like other animals, see the world through the lens of evolved adaptation}{The Evolution of Cognitive Bias, 2005 \cite{Haselton2005}}

Natural selection have shaped human perception, with the constraint of limited computational power. Cognitive biases led humans to draw inferences or adopt beliefs without corresponding empirical evidence. The fundamental work of Tversky and Kahneman \cite{Tversky1974} highlighted the existence of heuristics and systemic errors in judgment. However, those biases helped to solve adaptive problems, i.e. ``problems that recurred across many generations during a species' evolutionary history, and whose solution statistically promoted reproduction in ancestral environments'' \cite{Cosmides1994}.

\subsubsection{Limited rationality}

The economist Herbert A. Simon opposed the classical view of the \textit{rational} economic agent. He viewed humans as organisms with limited computational powers, and introduced the concept of \textit{bounded rationality} to take those limits into account in decision-making. \cite{Herbert1955}.

Cosmides and Tobby also criticized the study of economic agents as following "rational" decision rules, without studying the "computational devices" inside. Natural selection's invisible hand created the human mind, and economics is made of the interactions of those minds. Evolution led humans to develop domain-specific functions, rather than general-purpose problem-solvers. The intelligence of humans comes from those specific "reasoning instincts" that make inferences ``just as easy, effortless, and "natural" to humans as spinning a web is to a spider or building a dam is to a beaver'' \cite{Cosmides1994}.

\subsubsection{Heuristics}

The intractability of certain problems, the limited computational power of human minds, and uncertainty, are the most common ways to explain cognitive biases, and in particular heuristics, which are ``rules of thumb that are prone to breakdown in systematic ways'' \cite{Haselton2005}. Heuristics aim at reducing the cost of computing while delivering good-enough solutions. Processes are limited by brain ontogeny, i.e. the development of different parts of the brain, and complex algorithms take longer and require additional resources.

One of the most famous bias resulting from mental shortcuts is the "Linda Problem": individuals consider the assertion "Linda is a bank teller and active in the feminist movement" more probable than "Linda is a bank teller". This error is caused by the \textit{conjunction fallacy}, i.e. believing that the conjunction of two events is more probable than a single event \cite{Tversky1983}. Another notorious example is the Fundamental Attribution Error, attributing certain mental states to individuals because of their behavior, and not because of the logical implications of the context \cite{Jones1967} \cite{Ross1977}. 

\subsubsection{Error management biases}

Error Management Theory (EMT) studies cognitive biases in the context of error management. It distinguishes two error types \cite{Haselton2005}:

\begin{itemize}
    \item false positives (false belief)
    \item false negatives (failing to adopt a true belief)
\end{itemize}

One of the findings of EMT is that humans are biased to make the less costly error, even if it is the most frequent \cite{Haselton2006}. Biases of this kind include \cite{Haselton2005}:

\begin{itemize}
    \item \textit{Protective biases} (e.g. avoiding noninfectious person)
    \item \textit{Bias in Interpersonal Perception} (for instance sexual overperception for male and commitment skepticism for female)
    \item \textit{Positive Illusions} (estimates unrealistic likelihoods for positive events)
\end{itemize}

\subsubsection{Artifacts}

Humans might appear irrational in experiments because the tested abilities were not optimized by evolution. Those are called \textit{biases as artifacts} \cite{Haselton2005}. For instance, humans are better at statistical prediction if the inputs are presented in frequency form \cite{Gigerenzer1998}.

\section{Recommendations to Build a Safer AGI}

Humans have clear computational constraints (memory, processing, computing and clock speed) and have developed cognitive biases. An Artificial General Intelligence (AGI) is not \textit{a priori} constrained by such computational and cognitive limits.

Hence, if humans do not deliberately limit an AGI in its hardware and software, it could become a \textit{superintelligence}, i.e. "any intellect that greatly exceeds the cognitive performance of humans in virtually all domains of interest" \cite{Bostrom2014}, and humans could lose control over the AI. In this section, we discuss how to constrain an AGI to be less capable than an average person, or equally, while still exhibiting general intelligence. In order to achieve this, resources such as memory, clock speed, or electricity must be restricted. 

However, intelligence is not just about computing. Bostrom distinguishes three forms of superintelligence: speed superintelligence (``can do all that a human intellect can do, but much faster''), collective superintelligence (``A system composed of a large number of smaller intellects such that the system’s overall performance across many very general domains vastly outstrips that of any current cognitive system.'') and quality superintelligence (``A system that is at least as fast as a human mind
and vastly qualitatively smarter'') \cite{Bostrom2014}. A hardware-limited AI could be human-level-intelligent in speed, but still qualitatively superintelligent.

\subsection{Hardware}

To begin with, we focus on how to avoid speed superintelligence by limiting the AI's hardware. For instance, its maximum number of operations per second can be bounded by the maximum number of operations a human does. Similarly, by limiting its RAM (or anything that can be used as a working memory), we limit its processing power to process information at the similar rate as humans.

Focusing only on limiting the hardware is nonetheless insufficient. We assume that, in parallel, there exists other limitations (in software) that prevents the AI to become qualitatively superintelligent, to upgrade its hardware by changing its own physical structure, or to just buy computing power online.

\subsubsection{Storage Capacity}

\epigraph{ I should be surprised if more than $10^9$ was required
for satisfactory playing of the imitation game, at any rate against
a blind man. [...] A storage capacity of $10^7$
would be a very practicable possibility even by present techniques.}{Turing, Computing Machinery and Intelligence, 1950}

We estimated the storage capacity of the human brain to be at most $10^{15}$ bits, using one bit per synapse. The cost of hard drives have reached \$0.05/Gb in 2017 \cite{Backblaze2017}. Hence, the storage capacity of a human brain would cost at most approximately \$50,000. This is a pessimistic estimate: the brain uses maybe orders of magnitude less information for storage, and the price of a Gb could decrease even lower in the future.

To have a safe AGI, one should rather use much less storage capacity. For instance, as quoted in the epigraph, Turing \cite{Turing1950} estimated $10^{7}$ (or about 10Mb) to be a practical storage capacity to pass the Turing Test (and therefore attain AGI). Even if this seems very low, consider that an AGI could have a very elegant data structure and semantics, that may allow to store information much more concisely than our brains. In comparison, English Wikipedia in compressed text is about 12Gb, and is growing at a steady rate of 1Gb/year \cite{Wikipedia2018}. For this reason, allowing more than 10 Gb of storage capacity is unsafe. With 10Gb of storage, it could have access to an offline version of Wikipedia permanently, and be qualitatively superintelligent in the sense that it would have direct access to all human knowledge.

A counter-argument for such memory limit is that our brains process much more information than 10Mb when observing the world, and would store all those images in our long-term memory. The human-eye could observe, at most, 576Mb in a single glance (forgetting about all the visual flaws) \cite{Clarkvision2005}. However, all this resolution is not necessary to perform edge detection and image recognition. For instance, a 75x50 pixels image is enough to identify George W. Bush \cite{Yang2003}, and the popular database for handwritten digit recognition MNIST use 28x28 images \cite{lecun1998gradient}. Thus, we can imagine a "visual processing unit", that would transform the photons received by the captor of the AGI into a low resolution image, precise enough to be interpreted by our AGI, but still orders of magnitude smaller in size than a Mb.

\subsubsection{Memory access}

Manuel Blum opposes the application of traditional complexity theory to formalize how humans process information and do mental computations, in particular to generate a password from a private key previously memorized \cite{Blum2017}. In his Human-Model, memory can be modeled as a two-tape Turing machine: one for long-term memory, one for short-term memory. Blum considers potentially infinite tapes, because the size of the tape is not relevant for complexity theory, but for our purpose we can consider the tapes to be at most the size discussed previously for memory (e.g. 10Mb). According to Miller's magical number $7\pm2$ \cite{Miller1956}, human working-memory works with a limited amount of chuncks. So our two-tape Turing Machine should have a very short "short-term memory" tape, containing at most two or three 64-bit pointers pointing to chunks in the long-term memory (the other tape). More specifically, storing information in the long-term memory is slow, but reading from long-term memory (given the correct pointer) is fast. 

In modern computers, RAM's bandwidth is about 10GBytes/s, hard drive's bandwidth is 100MBytes/s, and with high clock rate a CPU can process about 25GBytes/s \cite{Lecoq2013}. In order to build a safer AGI, the memory access for the two mentioned tapes must be restricted, so that we are sure that data is being retrieved slower than by humans. However, the computing paradigms being very different, it is difficult to give a clear estimate of how much information is being retrieved per second by the brain \cite{Lecoq2013}.

\subsubsection{Processing}

We previously mentioned how the human brain could only process a limited amount of information per second. In addition to a limited number of chunks in working memory, other features must also be implemented to slow down the AGI and make it human-level intelligent. For instance, introduce some artificial period to process information, such as images, depending on the content. We already commented on the necessary duration of 100ms to process complex images \cite{Rousselet2004}. Similarly, the amount of time to process a certain image might depend on the complexity and size of the image. More generally, a model similar to \textit{Hick's Law} \cite{Hick1952} can be implemented, to have the AGI take linearly more time to take decisions as the information-theoretic entropy of the decision increases.

\subsubsection{Clock Speed}

As we mentioned multiple times, the brain parallelizes much more, using a totally different computing paradigm than the von Neumann architecture. Therefore, using a clock rate close to the frequency of the brain frequencies (typically \textasciitilde10Hz) is not relevant to our purpose, and it might prove difficult to build an AGI that exhibits human-level intelligence in real time using such a low clock rate.

To solve this, one possibility is to first measure better the trajectory of thoughts occuring in the brain, and then give a precise estimate of how frequently the processes in the brain are refreshed (i.e. evaluating some kind of clock rate). Another solution is to abandon the von Neumann architecture, and build the AGI with a computer architecture more similar to the human brain.

\subsubsection{Computing}

In "The Cognitive Limits of the Human Brain", we mentioned Bostrom's estimate \cite{Bostrom1998} of at most $10^{17}$ operations per second for the brain. This is a very large number, and could only happen if the AGI's hardware allowed that much computing power. This will not be the case, according to what we said previously in \textit{Storage capacity} and \textit{Memory access}.

More importantly, even if we could measure a number of operations per second that would actually be lower than any number of operations per second a human brain does for any given task, it might not be a correct bound. Why? The brain has evolved to achieve some very specific tasks, useful for evolution, but nothing guarantees that the complexity or the processes happening in the brain are algorithmically optimal. Thus, the AGI could possess a structure that would be far more optimized for computing than the human brain.

Therefore, restricting the number of operations alone is insufficient: the algorithmic processes and the structure of the AGI must be precisely defined so it is clear that the resulting processes happening are performing tasks at a lower rate than humans.

\subsection{Software}

In November 2017, there were more than 45 companies in the world working on Artificial General Intelligence \cite{Baum2017}. Ben Goertzel distinguishes three major approaches \cite{Goertzel2018}:

\begin{itemize}
    \item 1) Use neural networks to emulate the different parts of the brains, visual and auditory processing, and connect all those parts together by emulating how those parts talk to each other. Deepmind is a representative example.
    \item 2) Take Marcus Hutter's Universal Artificial Intelligence model \cite{hutter2004universal} and try to limit the required computing power
    \item 3) Ben Goertzel's approach with OpenCog: look at the cognitive processes happening in the brain from a high-level cognitive point of view, and map this into a weighted labeled hypergraph.
\end{itemize}

In this paper, in order to build a safe AGI with at most human intelligence, we will focus on the first approach. Indeed, a more universal or high-level Artificial General Intelligence will have a very different computing paradigm than the human brain, so it would be difficult to restrain the AGI computing resources accordingly.

In addition to those neural networks emulating processes happening in the brain, we consider additional safety measures that must be implemented to obtain a safe AGI.

\subsubsection{No self-improvement}

A limited initial hardware is not a real restriction if the AGI can buy some additional computing power online or change its own structure. To prevent the AGI from changing its own code, one possibility is to hard-code the feature directly "you shall not rewrite your own code". Another (subtler) possibility is to encrypt its source code, making self-modification more difficult. 

However, it might seem that this does not completely solve the problem, as the AGI could manipulate humans into changing its code. Yet, with our AGI design, the AGI would not be superintelligent, but at most human-intelligent. So the AGI would not have the necessary "social manipulation superpowers" \cite{Bostrom2014} to convince humans to change its code.

\subsubsection{Cognitive Biases}

Humans have developed cognitive biases because of natural selection. Incorporating human biases into the AGI present several advantages: they can limit the AGI's intelligence and make the AGI fundamentally safer by avoiding behaviors that might harm humans. Thus, heuristics limiting the possible results of a computation, or error management biases, can help build a less capable AGI that would also make less errors.

Here is a list of cognitive biases that could make the AGI safer \cite{List2018}:

\begin{itemize}
    \item \underline{Planning fallacy}: would prohibit the AGI from successfully planning a takeover or a treacherous turn
    
    \item \underline{Bandwagon Effect}: the AGI will acquire human values that are shared among the group it belongs to
    
    \item \underline{Confirmation Bias}: the AGI will rationalize and confirm that it is useful to help humans, or that AI Safety is an important problem
    \item \underline{Conservatism}: the AGI will keep the same initial values, and not become evil
    \item \underline{Courtesy bias}: the AGI will try to not offend anyone, avoiding aggressive behaviors
    \item \underline{Functional Fixedness}: the AGI will only use objects like humans do. It will not “hack” anything or use objects with malicious intent
    \item \underline{Information bias}: the AGI will have the tendency to seek information and think more, avoiding errors
    \item \underline{Mere-exposure effect}: the AGI will have good intention towards humans because it will be exposed to humans
    \item \underline{Omission bias}:  bad actions will be much lower in its utility function than status quo
    \item \underline{Status quo bias}: an action that does not change the environment or even  inaction will be desirable
    \item \underline{Authority bias}: some human values are shared among most authority figures. The AGI will avoid extreme individual opinions and prefer conventional ones
    \item \underline{System justification}: the AGI will defend human values, and won't promote revolution
    \item \underline{Processing difficulty effect}: the AGI will prefer well-thought arguments and well-argued ideas over simple actions or decisions that can be harmful
    \item \underline{Spotlight effect}: the AGI will overestimate the amount of people observing its behavior. Hence, it will behave in a similar fashion  both with high and low supervision
\end{itemize}

It is clear that some of those biases can lead to irrational behavior. Therefore, even though those biases can lead to safer behavior, they must be implemented with caution.

\subsubsection{Avoiding a treacherous turn}

In the Chapter 8 of \textit{Superintelligence} \cite{Bostrom2014}, Bostrom notes the following: behaving nicely in a sandbox environment is a convergent instrumental goal, both for a friendly or a malicious AI. At one point, an AI could exhibit enough intelligence to understand that it is being observed and could restrain its own intelligence accordingly. Bostrom calls this first moment of realization ``the conception of deception'', and would be the last moment of vulnerability where we could observe a true malicious intent. After this point, the behaviors will appear "safe", until a certain threshold is reached, where the AI does not feel threatened anymore by its supervisors, and decides to take a treacherous turn and reveal its true nature by taking over.

In the implementation of the AI, however, we can decide to make it very clear what processes are happening, when, and how much computing power it took to do each task. This interface would allow the human supervisors to better monitor the behavior of the AI in the sandbox environment, and detect a treacherous turn. Additionally, the designers can implement a safety feature that consists in preventing the AI from erasing any of the log files, or from even knowing that it is being observed.

\section*{Conclusion}

In the history of Artificial Intelligence, one of the greatest challenge has been to pass the Turing Test. To win at the Imitation Game, chatbots are made \textit{Artificially Stupid}. More generally, introducing Artificial Stupidity into an AI can improve its interaction with humans (for instance in video-game), but can also be a safety measure to better control an AGI. In this paper, we proposed a design of a safer and humanly-manageable AGI. It would be hardware constrained, so it would have less computing power than humans, but also an architecture very similar to human brains. Additionally, some features in software might help avoid self-improvement, a treacherous turn, or just make the AI safer (e.g. cognitive biases).

This approach presents several limitations and makes multiple assumptions:
\begin{itemize}
    \item this paper is limited to the case where the first approach to AGI (emulating a human-brain with neural networks) is possible, and will be the first approach to succeed
    \item prohibiting the AGI from hardware or software self-improvement might prove a very difficult problem to solve, and may even be incompatible with corrigibility
    \item  it may be impossible to build an AI that simultaneously operates under such heavy constraints and still exhibits general intelligence
    \item this approach does not generalize: it is impossible to build a safe Superintelligence from this AGI design
\end{itemize}

Therefore, progress must still be made to generalize this approach. Future directions of this research include:
\begin{itemize}
    \item exploring other limits of human cognition
    \item determining how to apply those limits for different AGI designs (Marcus Hutter's approach or Ben Goertzel's approach for instance)
\end{itemize}

\printbibliography

\end{document}